\documentclass[11pt]{article}
\usepackage[margin=1in]{geometry}
\usepackage{newtxtext,newtxmath}
\usepackage{booktabs}
\usepackage{multirow}
\usepackage{graphicx}
\usepackage[hidelinks]{hyperref}
\usepackage[numbers,sort&compress]{natbib}
\usepackage{xcolor}
\usepackage{caption}
\newcommand{\mortal}{Mortal}

\title{\bf Text Is the Bottleneck:\\ A Small-Scale Replication of LLM Game Playing in Riichi Mahjong}
\author{Mingge Xu\thanks{Department of Bioengineering, Graduate School of Engineering, The University of Tokyo. Email: \texttt{xu-mingge580@g.ecc.u-tokyo.ac.jp}}}
\date{October 2026}

\begin{document}
\maketitle

\begin{abstract}
We replicate the riichi-mahjong track of \emph{Can Large Language Models Master Complex Card Games?} \citep{llm4cardgame} under a drastically smaller budget: models of at most 2B parameters, a single 16\,GB consumer GPU, about five GPU-days.
\textbf{(1) With matched data and labels (LLMs LoRA-tuned for one epoch), text is the bottleneck.} On the same 100k-decision training set, a 2.6M-parameter convolutional \emph{board encoder} agrees with the strong AI \mortal{} on 80.6\% of test decisions, above every text-prompt LLM we train (best 77.8\%), and places better at the table (2.85 over 1{,}000 games vs.\ $\approx$3.0 for the LLMs; paired difference at $n{=}100$ not significant). Scaled to 1.71M decisions and distilled from \mortal{}'s Q-values, a 4.9M-parameter Transformer reaches 94.7\% agreement, beats the 0.8B LLMs on identical seeds (sign test, $-0.45$ ranks, $p{=}0.005$), and is statistically \emph{equivalent} to parity against three \mortal{} champions (TOST, $\pm0.10$ ranks, $p{=}6{\times}10^{-4}$)---noting that parity against a teacher-opponent is roughly the ceiling of pure imitation.
\textbf{(2) Training toward the teacher raises teacher agreement more than a 2.5$\times$ parameter increase does} ($+2.5$ for labels at 0.8B vs.\ $+1.5$ for 0.8B$\to$2B under human labels, McNemar $p\approx2\times10^{-9}$ / $0.012$); on human-record agreement both effects are $+1.0$, and neither is visible at the table.
\textbf{(3) Few-point agreement differences do not resolve into playing strength within a family.} Across families the ordering is clean (scaled encoder: 94.7\% agreement, placement 2.51, TOST-equivalent to parity; matched-data encoder 80.6\%, 2.85; LLMs 74.3--77.8\%, 2.95--3.20 across blocks)---though only the scaled encoder separates significantly, and it also has 17$\times$ more data. Within the text-LLM family, $+2.5$ agreement points coincide with a paired placement difference of $+0.02\pm0.12$ ranks (sign test $p{=}1.0$) and $+2.0$ points at 2B with $-0.06$ ($p{=}0.37$)---both far below the $\approx$0.25-rank resolution of 100 paired games.
We also document a paper--code mismatch in the original ``winner filter'', reconstruct its two private engine patches, and show that a divide-and-conquer pipeline (encoder for decisions, program-supplied facts for the rationale) makes template-generated Japanese rationales 93\% sentence-exact (43\% without facts)---a measure of faithful integration of program-supplied facts, since a template copier would also ace the checkable categories. Per Tenhou's terms we release no trained weights---only code, configurations, and evaluation protocols.
\end{abstract}

\section{Introduction}
\label{sec:intro}

\citet{llm4cardgame} trained 7B--9B instruction-tuned LLMs with LoRA \citep{lora} on text descriptions of eight card games, including four-player riichi mahjong, and conclude that ``LLMs can approach the performance of strong game AIs through supervised fine-tuning on high-quality data.'' The mahjong track was trained on 8$\times$H100 with expert game records from Tenhou, and evaluated by playing against \mortal{} \citep{mortal} (50 games per agent, scored 3/2/1/0 by placement; the best mahjong model reaches $\approx$1.4 reward, $\approx$2.6 average placement, against their own---unreleased---\mortal{} checkpoint).

This paper asks what survives of that conclusion when the budget shrinks by roughly two orders of magnitude: one RTX 5070 Ti (16\,GB), models up to 2B parameters (Qwen3.5-0.8B/2B, Qwen2.5-0.5B, TinySwallow-1.5B), and LoRA rather than full fine-tuning. Small-scale replications are useful twice over: they test whether the claimed effect needs scale, and they isolate representation and supervision effects that large runs smooth over.

Our contributions:
\begin{itemize}\itemsep2pt
\item \textbf{A faithful replication} of the mahjong data pipeline, including a reconstruction of the two engine patches the original code omits (\S\ref{sec:setup}), with per-game decision counts matching the paper (654.6 vs.\ 656.92).
\item \textbf{Representation results, offline and at the table} (\S\ref{sec:offline},~\S\ref{sec:arena}): on identical data and labels, a tiny board encoder beats all text-prompt LLMs at imitating \mortal{} (80.6\% vs.\ 77.8\%) and is nominally better at the table (2.85 vs.\ 2.95--3.03; paired difference not significant at $n{=}100$); the scaled, distilled Transformer encoder is statistically equivalent to parity against three \mortal{} champions (2.51$\pm$0.03 over 1{,}000 games, TOST) and beats the 0.8B LLMs on identical seeds ($p{=}0.005$).
\item \textbf{Supervision results} (\S\ref{sec:offline}): teacher-label training raises teacher agreement more than a 2.5$\times$ parameter increase does; on human-record agreement the two effects are about equal.
\item \textbf{The game-play evaluation} the original protocol prescribes (\S\ref{sec:arena}): trained LLMs reach $\approx$3.0 average placement against three \mortal{} agents---far above a random-legal floor (3.90) and below parity (2.50)---while the scaled encoder reaches parity (TOST-equivalent over 1{,}000 games). Agreement with \mortal{} orders the agents across representation families; \emph{within} the text-LLM family, few-point agreement differences do not resolve into placement differences at 100-game power.
\item \textbf{A divide-and-conquer pipeline} (\S\ref{sec:pipeline}): freezing the encoder and letting a program supply checkable facts makes the LLM's Japanese-language rationales 93\% sentence-exact (43\% without facts; a template copier would also ace the checkable categories); a perception module extends the same pipeline from pixels (92.7\% agreement with \mortal{}).
\item \textbf{Replication notes and negative results} (\S\ref{sec:replnotes},~\S\ref{sec:negative}), including a paper--code mismatch in the original winner filter and several silent-failure pitfalls of modern tooling.
\end{itemize}

We release the full evaluation harness---including the reconstructed arena integration that the original artifact omits---alongside every arena game log.

\section{Background and Related Work}
\label{sec:related}

\paragraph{LLMs playing cards.} \citet{llm4cardgame} is the direct target of our replication; its mahjong track uses text prompts of 36 state fields and legal-action lists, SFT on (state, expert action) pairs, and evaluation vs.\ \mortal{}. Related game AIs built by RL include DouZero for DouDizhu \citep{douzero}, Suphx for mahjong \citep{suphx}, and \mortal{} itself \citep{mortal}, which provides both our teacher signal and our arena opponent; other open mahjong-AI stacks include kanachan \citep{kanachan} and the mjx simulator \citep{mjx}. \citet{ruoss2024amortized} reach grandmaster-level chess with a from-scratch transformer over a compact FEN tokenization---closer in spirit to our board encoder than to a pretrained LLM reading prose. The open contrast is compact-encoding-plus-from-scratch-model versus prose-prompt-plus-pretrained-LLM, and at what parameter scale the latter catches up. Suphx combines supervised pretraining with RL \citep{suphx}, as does \mortal{} itself \citep{mortal}---which is why, in \S\ref{sec:arena}, we read parity against \mortal{} as the ceiling of pure imitation.

\paragraph{Riichi mahjong.} A four-player, imperfect-information game with 136 tiles and rich conventions (riichi declarations, dora, furiten). A full hanchan lasts $\approx$650--660 decisions in the LLM4CardGame encoding.

\paragraph{Mahjong AIs as teachers.} \mortal{} is a distributed RL agent over the \texttt{libriichi} engine. We use a community checkpoint (192 channels $\times$ 40 blocks, 10.8M parameters, trained 298k steps) \citep{voidshine298k}. Tenhou phoenix-table (houou) players agree with \mortal{}'s first choice on 81.1\% of 153{,}244 logged decisions and place the human action in its top two 94.73\% of the time (81.9\% top-1 on our 5k test set; the top-2 figure coincides numerically with the TF encoder's top-1 in Table~\ref{tab:offline}, but they are unrelated quantities). We use \mortal{} as teacher and arena opponent because VoidShine-298k is, to our knowledge, the strongest publicly available \mortal{} checkpoint; we have no independent rating for it.

\section{Replication Setup}
\label{sec:setup}

\paragraph{Data.} We use a 2026 dump of Tenhou houou (top-tier) games, 12{,}188 hanchan in total.
Games 1--7{,}000 form the training pool (1{,}086{,}758 SFT decisions after the released code's filtering), games 7{,}001--8{,}000 are held out, and 5{,}000 decisions sampled from them (fixed seed, after the same winner filter---so \emph{every} agreement number here, including the 81.9\% overlap, is computed on the decisions of hand-leading players; the 81.1\% figure in \S\ref{sec:related} is unfiltered) form the \emph{unified test set} on which every offline number in this paper is computed. Games 8{,}001--12{,}188 (655{,}557 decisions) are added only for the best encoders: 1{,}742{,}315 raw, 1{,}708{,}593 after the validation split (3\% of \emph{games}, 1.9\% of decisions) (the ``1.09M'' CNN set likewise counts 1{,}086{,}758 raw).
The training subset for LLM experiments is a fixed 100k-decision sample; 1k of it is split off for validation.

\paragraph{Prompt and training.} We reuse the original prompt verbatim (rules, 36 state fields, legal-action list, JSON output instruction). Models are Qwen2.5-0.5B \citep{qwen25}, TinySwallow-1.5B \citep{tinyswallow}, and Qwen3.5-0.8B/2B---successors of Qwen3 \citep{qwen3}; no technical report at the time of writing, so we cite the model cards.\footnote{\url{https://huggingface.co/Qwen/Qwen3.5-0.8B}, \url{https://huggingface.co/Qwen/Qwen3.5-2B}; accessed 2026-10-25} All LLMs are trained with LoRA \citep{lora} (rank 8 on all linear layers: 5.4M trainable parameters for the 0.8B, 8.4M for the 2B---the same order as the encoders they compete with---1 epoch, lr $10^{-4}$ cosine, effective batch 16) using LLaMA-Factory \citep{llamafactory}; no label smoothing. A 0.8B model trains in 9.8\,h; a 2B model in 13.1\,h, on the single GPU (the one exception in the whole project: the 2B \mortal{}-label run, whose training briefly exceeded local memory, was finished on a rented A100-40\,GB for \$21; all evaluation is local).

\paragraph{Reconstructing the missing engine patches.} The released LLM4CardGame code calls two \texttt{libriichi} extensions that only exist in the authors' private fork: an observation extractor (\texttt{take\_obs\_raw}) on the data side, and a \texttt{mortal2} engine type that lets a Python LLM agent drive one seat of the \mortal{} arena on the evaluation side. We re-implemented both against the open \texttt{libriichi} and verified behavior: per-game decision counts 654.6 vs.\ the paper's 656.92, and a stub-agent smoke test exercising all action paths (including kan double-asks).

\paragraph{Metrics.} \emph{Offline agreement}: fraction of test decisions whose predicted action equals the reference (human record, or \mortal{}'s first choice). \emph{Game play}: the original protocol---one challenger seat vs.\ three \mortal{} champions, seat-rotated seeds, placement points $90/45/0/-135$ (\S\ref{sec:arena}).

\section{Offline: Representation, Labels, and Scale}
\label{sec:offline}

\begin{table}[t]
\centering\small
\caption{Offline results on the unified 5{,}000-decision test set. Agreement is with the human record (H) or with \mortal{}'s first choice (M). All LLMs use text prompts; the encoder consumes a $58\times34$ plane + 71-scalar tensor computed from the same prompt. ``Hard'' = \mortal{}'s first choice only (identical supervision to the text LLMs, plus label smoothing 0.05); ``soft'' = additionally distilling \mortal{}'s Q-value distribution. TinySwallow is trained on 20k decisions (memory-limited) and listed for reference.}
\label{tab:offline}
\begin{tabular}{llcc}
\toprule
Model & Supervision & H (\%) & M (\%) \\
\midrule
human record (reference) & --- & 100 & 81.9 \\
\midrule
\multicolumn{4}{l}{\emph{Text-prompt LLMs (LoRA, 100k decisions unless noted)}} \\
Qwen2.5-0.5B & human & 68.5 & 71.4 \\
TinySwallow-1.5B (20k) & human & 59.9 & 61.1 \\
Qwen3.5-0.8B & human & 70.6 & 74.3 \\
Qwen3.5-0.8B & \mortal{} labels & 71.6 & 76.8 \\
Qwen3.5-2B & human & 71.6 & 75.8 \\
Qwen3.5-2B & \mortal{} labels & 72.2 & \textbf{77.8} \\
\midrule
\multicolumn{4}{l}{\emph{Board encoders (no LLM)}} \\
CNN 2.6M, 100k decisions & \mortal{} hard & 74.0 & 80.6 \\
CNN 2.6M, 1.09M & \mortal{} soft & 80.5 & 92.2 \\
Transformer 4.9M, 1.71M & \mortal{} hard & 80.7 & 91.6 \\
Transformer 4.9M, 1.71M & \mortal{} soft & \textbf{81.9} & \textbf{94.7} \\
\bottomrule
\end{tabular}
\end{table}

\begin{figure}[t]
\centering
\includegraphics[width=0.78\linewidth]{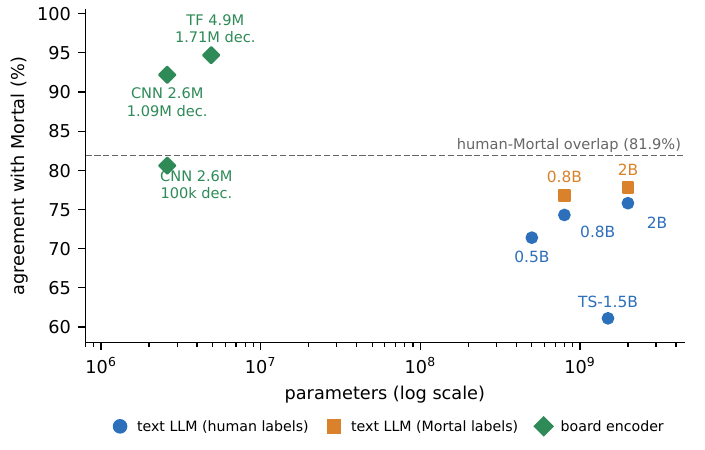}
\caption{Agreement with \mortal{}'s first choice vs.\ parameter count (log scale). Text-prompt LLMs (circles: human labels; squares: \mortal{} labels) reach 71--78\% across the 0.5B--2B range (TS-1.5B, 61\%, was trained on 20k decisions---its 100k run was interrupted at step 5{,}005/6{,}188 and is not evaluated), while board encoders (diamonds) consuming the same prompt fields as tensors reach 80.6--94.7\%; the dashed line marks human houou experts (81.9\% on this test set). Encoder annotations give training-set size in decisions.}
\label{fig:bottleneck}
\end{figure}

\paragraph{Text is the bottleneck---with matched data and labels.} Table~\ref{tab:offline} and Figure~\ref{fig:bottleneck} show the central result. On the same 100k decisions, the 2.6M-parameter CNN encoder reaches 80.6\% agreement with \mortal{}---2.8 points above the best text LLM (77.8\%), which has 310--770$\times$ the \emph{total} parameters (trainable: 5.4--8.4M vs.\ 2.6M) and trains $\sim$370$\times$ longer in wall clock (9.8\,h for the 0.8B, 13.1\,h for the 2B, vs.\ 1.6\,min for the encoder). Three caveats scope the claim. (i) \emph{Supervision is matched}: the encoder's ``hard'' row uses \mortal{}'s first choice only, exactly the labels the text LLMs see; the encoders additionally use label smoothing 0.05 over the 46-action space while the LLMs use none. (ii) \emph{Optimization is not matched}: the encoder trains all parameters for up to 16 epochs with early stopping, while each LLM is tuned once with LoRA rank 8 ($\approx$10M trainable parameters) for one epoch; we did not run multi-epoch or full fine-tuning ablations, so the claim is about \emph{text prompting with matched data and labels under one-epoch LoRA}, not about the architectures in general; the output heads also differ (46-way classifier with a legal mask vs.\ free-form JSON generation), which we cannot separate from the interface. (iii) \emph{Information is matched by construction}: every plane and scalar is a deterministic function of the same 36 prompt fields (the featurizer consumes the prompt; quantities the prompt already reports, such as shanten and waits, are re-encoded, not invented), so the gap measures what serializing those fields into $\sim$1{,}310 tokens of prose costs a learner at this scale. At larger data scales the encoders additionally distill \mortal{}'s Q-distribution (``soft''); a hard-label control trained on the same 1.71M decisions reaches 91.6\%, so the label mixture accounts for only the last 3 points of the 94.7\% offline (the arena uses the distilled model; the hard control was not played). Scaling encoder data and switching to a small Transformer reaches 94.7\% agreement---higher than the 81.9\% \emph{overlap} between human experts and \mortal{} (a behavioral coincidence rate, not a human ceiling)---in 28 minutes of training.

\paragraph{Teacher labels raise teacher agreement---more than scale does.} Reading the $2{\times}2$ of Table~\ref{tab:offline} on the \mortal{}-agreement column: at 0.8B, human$\to$\mortal{} labels give $+2.50$ points ($p\approx2\times10^{-9}$, McNemar, treating test decisions as exchangeable---they cluster mildly within games); under human labels, 0.8B$\to$2B gives $+1.50$ ($p{=}0.012$); under \mortal{} labels the scale effect is $+1.00$; both changes together give $+3.50$ ($p\approx1\times10^{-13}$). The label effect is partly definitional---training toward \mortal{}'s action raises agreement \emph{with \mortal{}}---and on the human-record column the picture flattens: the label effect at 0.8B is $+1.0$ and the scale effect under human labels also $+1.0$. Neither effect is visible at the table (\S\ref{sec:arena}), and each configuration is trained once, so single-run training noise (unmeasured) applies to all cells.

\paragraph{Zero-shot.} The untrained Qwen3.5-0.8B base never emits valid action JSON (0\% parseable within 64 or even 512 tokens); the untrained 2B base does emit valid JSON but agrees with the record on only 13.1\% of decisions (below a first-legal-action policy's 14.4\%). Post-training is what makes the small models operational at all.

\section{Playing the Game: Arena Evaluation}
\label{sec:arena}

The original paper's headline metric is game play against \mortal{}; their reported numbers (50 games per agent, 3/2/1/0 reward, best $\approx$1.4 reward $=$ $\approx$2.6 average placement---the reward is an affine transform of placement, so the \emph{metric} is comparable) differ from ours in opponent checkpoint (theirs is unreleased) and game count, so we compare ranges, not points. We implement the protocol with the reconstructed \texttt{mortal2} arena, decoding greedily (the paper's own configuration) with the \emph{exact} chat template used in training. The board encoders play through the same interface, replacing the LLM with a single forward pass. Every LLM query across the campaign (74{,}535 in total) returned parseable, legal JSON (the encoders are legal by construction).

\begin{table}[t]
\centering\small
\caption{Game-play results: one agent vs.\ three \mortal{} (VoidShine-298k) champions. 100 games each (25 seeds $\times$ 4 seat rotations), greedy decoding. Placement distribution $[1^{\text{st}},2^{\text{nd}},3^{\text{rd}},4^{\text{th}}]$, average placement $\pm$ s.e., average Tenhou-style points ($90/45/0/-135$). Parity with the champion pool is 2.50. Our primary metric is \emph{average placement} (the original paper's protocol); the points column can rank differently because a 4th place costs $-135$ (e.g., the 2B model's extra first places do not offset its extra last places). Encoder rows are 1{,}000 games (250 seeds $\times$ 4 rotations, distributions as percentages) on the disjoint 95k seed block with seed-clustered s.e.; all other rows are 100 games. The encoders' 100-game blocks (91k/92k) give 2.55/2.56 (TF) and 2.90/2.96 (CNN), consistent with the 1{,}000-game values.}
\label{tab:arena}
\begin{tabular}{lcccc}
\toprule
Agent & Placement dist. & Avg.\ rank & 1st rate & Avg.\ pt \\
\midrule
random legal action & $[0,1,8,91]$ & 3.90$\pm$0.03 & 0\% & $-122$ \\
\midrule
Qwen3.5-0.8B, human labels & $[10,16,35,39]$ & 3.03$\pm$0.10 & 10\% & $-36$ \\
Qwen3.5-0.8B, \mortal{} labels (block A) & $[7,18,23,52]$ & 3.20$\pm$0.10 & 7\% & $-56$ \\
Qwen3.5-0.8B, \mortal{} labels (block B, paired w.\ above) & $[10,23,23,44]$ & 3.01$\pm$0.10 & 10\% & $-40$ \\
Qwen3.5-2B, human labels & $[14,16,25,45]$ & 3.01$\pm$0.11 & 14\% & $-41$ \\
Qwen3.5-2B, \mortal{} labels (paired w.\ 2B-human) & $[14,17,29,40]$ & 2.95$\pm$0.11 & 14\% & $-34$ \\
\midrule
CNN encoder 2.6M (100k dec.), 1{,}000 games & $[16.9,19.9,24.1,39.1]$ & 2.85$\pm$0.03 & 17\% & $-29$ \\
Transformer enc.\ 4.9M (1.71M dec.), 1{,}000 games & $[25.6,24.3,23.6,26.5]$ & \textbf{2.51}$\pm$0.03 & \textbf{26\%} & $\mathbf{-2}$ \\
\bottomrule
\end{tabular}
\end{table}

\begin{figure}[t]
\centering
\includegraphics[width=\linewidth]{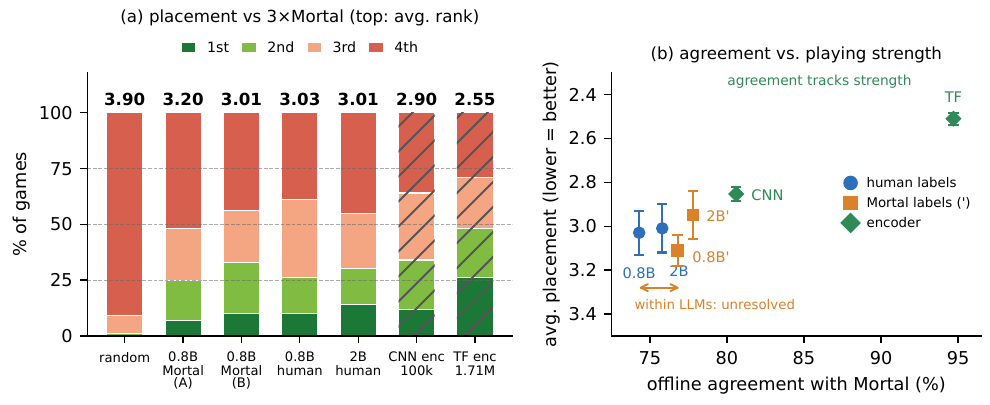}
\caption{Game-play evaluation vs.\ three \mortal{} champions (100 games per bar unless noted). (a)~Placement distributions with average ranks above the bars; hatched bars are board encoders; dashed lines mark the cumulative placement boundaries under parity (25/50/75\%). (b)~Offline \mortal{}-agreement vs.\ average placement (error bars: $\pm$1 s.e.; H = human labels, M = \mortal{} labels pooled over two seed blocks): \emph{across} representation families agreement tracks strength; \emph{within} the text-LLM family few-point agreement differences do not resolve into placement differences (paired $+0.02\pm0.12$ ranks for $+2.5$ agreement points).}
\label{fig:arena}
\end{figure}

Four observations. First, \textbf{the trained small LLMs genuinely play}: average placement $\approx$3.0 is far from the random floor (3.90); they finish 1st or 2nd in 25--33\% of games against three copies of a strong opponent.
Second, \textbf{the scaled encoder is statistically equivalent to parity, and beats the LLMs on identical seeds}. Over 1{,}000 games the Transformer encoder averages $2.511\pm0.028$ (seed-clustered s.e.; 95\% CI $[2.46,2.57]$) vs.\ parity 2.50---a two-one-sided-tests (TOST) equivalence test with margin $\pm0.10$ ranks gives $p{=}6{\times}10^{-4}$, i.e.\ \emph{equivalence}, not merely failure to reject (the margin was fixed before reading the results, at roughly the between-block s.d.\ of a difference, $\approx$0.14; the paired-power note: 100 paired games resolve $\approx$0.25-rank effects at 50\% power, $\approx$0.34 at 80\%). Two framing caveats: the opponent is also the teacher, and a perfect clone seated against three copies of itself scores 2.50 by symmetry, so parity here is roughly the \emph{ceiling of pure imitation} rather than an independent strength benchmark; and the scaled encoder sees 17$\times$ more decisions than the LLMs, so this row is not a matched comparison. The matched-data picture: CNN encoder 2.85--2.96 over 1{,}200 games vs.\ LLMs 2.95--3.03, paired differences against the two 0.8B models $-0.05/-0.03$ (per-game log recomputation; sign tests $p{=}0.6/0.8$: within about half a standard error); on the same 92k seeds the TF encoder beats both 0.8B models by $-0.47/-0.45$ ranks ($p{=}0.005$), and on the 91k block it beats the matched-data CNN by $-0.34$ ($p{=}0.02$). All of this is cheap: 1{,}000 encoder games take 19--23 minutes, versus 1.5--4 GPU-hours per 100 LLM games (0.5\,s per decision).
Third, \textbf{all five LLM blocks (four configurations) remain clearly below parity} (2.50): text-prompt LLMs at this scale are substantially weaker than \mortal{}, consistent with the offline gap (74--78\% agreement vs.\ 81--95\% for the encoders). Pooling the two \mortal{}-labeled blocks gives $3.11\pm0.07$ ($n{=}200$), within 0.1 ranks of the human-labeled models' 3.01--3.03. Because the arena opponent is the teacher, we add two checks with other opponents: against a pool of the weaker matched-data encoders the TF encoder wins clearly ($2.26\pm0.11$), and against an alternative opponent---three copies of a small network we distilled ourselves from \mortal{}-298k's Q-values (weaker than 298k, used only as a sparring partner)---it is indistinguishable from parity ($2.48\pm0.12$).
Fourth: \textbf{within the text-LLM family, few-point agreement differences do not resolve into placement differences} (Figure~\ref{fig:arena}b). On 100 exactly paired games (identical seeds and seats, 92k block) the two 0.8B models split 29--30 with 41 ties (29 wins for the \mortal{}-labeled model, 30 for the human-labeled): mean rank difference $+0.02\pm0.12$ under the convention difference $=$ \mortal{}-labeled minus human-labeled (sign test $p{=}1.0$). The 95\% CI ($\approx[-0.22,+0.26]$) is wider than the $\approx$0.1-rank effect the cross-family trend would predict for a 2.5-point gap---100 paired games cannot resolve differences below $\approx$0.25 ranks, so the correct reading is \emph{undetectable}, not \emph{zero}. The 2B pair (94k block, same convention) replicates this: $+2.0$ agreement points coincide with $-0.06\pm0.13$ paired ranks ($p{=}0.37$), with the \mortal{}-labeled 2B (2.95) nominally the best LLM at the table. Our first campaign ran the 0.8B pair on \emph{disjoint} seed blocks and we initially read a 0.17-rank inversion (3.03 vs.\ 3.20) into it---about 1.2 s.e., consistent with ordinary sampling noise ($\approx$0.14-rank between-block s.d.), and a lesson about single-block evaluations that we took literally. What the paired logs \emph{do} resolve are style differences: the \mortal{}-labeled 0.8B melds far more (3.3 vs.\ 2.2 per game; paired Wilcoxon on per-game counts, $p<10^{-3}$) and wins more (1.81 vs.\ 1.55; $p{=}0.02$), with a higher but individually non-significant deal-in rate (80\% vs.\ 75\% of games). We read this as an aggressive tempo style; whether it stems from the teacher's labels or from single-run training noise, the paired design cannot say. No multiple-comparison correction is applied to the many tests in this section; they are descriptive.

\paragraph{Cost of evaluation.} Each 100-game LLM run costs $\approx$1.5--4 GPU-hours (0.5\,s per decision, $\approx$169 decisions per seat-game); 1{,}000 encoder games take 19--23 minutes. The full campaign---five LLM blocks plus a random floor, four encoder 100-game blocks, two encoder 1{,}000-game runs, two extra-opponent matches, and a 32-game symmetry check, $\approx$3{,}230 games and 74{,}535 LLM queries in total (inventory in Appendix~\ref{app:arena})---costs under one GPU-day: the prescribed evaluation is cheap enough not to skip.

\section{Divide-and-Conquer: Structured State, Program Facts, Pixels}
\label{sec:pipeline}

\begin{figure}[t]
\centering
\includegraphics[width=\linewidth]{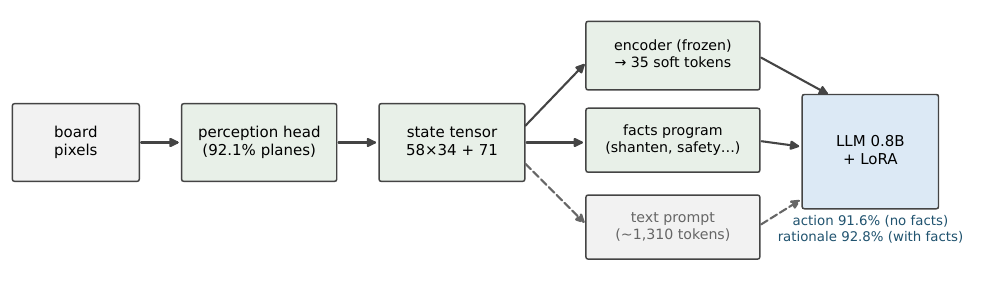}
\caption{The divide-and-conquer pipeline. A small perception head turns pixels into the structured state; a frozen CNN encoder compresses it into 35 soft tokens; a program appends checkable facts (including the encoder's ranking in the rationale setting); the LLM integrates them. The percentages shown are from the clean-state pipeline: action 91.6\% without facts, rationale sentences 92.8\% with facts. The dashed path is the original paper's text-only baseline.}
\label{fig:pipeline}
\end{figure}

The representation result motivates replacing the text prompt with structure, while keeping the LLM for what it is good at (Figure~\ref{fig:pipeline}).

\paragraph{Encoder tokens.} We encode the position with the (frozen) CNN encoder trained on 1.09M decisions---the 92.2\% M-agreement checkpoint of Table~\ref{tab:offline}, whose outputs are 34 per-tile vectors plus 1 global vector---and inject these 35 vectors as soft tokens into the same Qwen3.5-0.8B via a 1.3M-parameter projector, training only the projector+LoRA on \emph{Japanese rationales + actions}. Without any auxiliary facts the model reaches 91.6\% \mortal{}-agreement---0.6 points \emph{below} its frozen encoder (92.2\%), i.e.\ the LLM learns to read off the encoder's decision but adds nothing.

\paragraph{Program facts.} We then let a program append checkable facts (acceptances/shanten, safety, score gaps) to the rationale target; the rationale targets themselves are template-generated Japanese captions built from the game records, and ``sentence precision'' is an exact string match per sentence against the program's ground truth. Rationale sentence precision rises 43.1\%$\to$88.7\% (candidate lists only) $\to$92.8\% (with the \emph{encoder's} value-head ranking of the top actions appended---the same ranking that feeds the action, which is why the action then equals the encoder's by construction); every sentence category that is programmatically checkable (tile efficiency, situation, calls, safety, riichi) reaches \textbf{100\%}, and the one imperfect category (79.5\%) is the teacher-comparison phrasing, whose \emph{identity} is given by the ranking but whose \emph{magnitude} wording (``narrowly better than\ldots'') is not. Without facts the model hallucinates numbers (``47 acceptances'' when there are 46); with facts, the checkable sentences are correct by construction of the metric---a template copier would also score 100\% there---so the quantity measured is \emph{faithful integration} (the model binds the right numbers to the right tiles), not judgment. The 92.8\% claim is about the \emph{rationale}; the LLM's own decision ability without the ranking is the 91.6\% figure.

\paragraph{From pixels.} Rendering the board (synthetic, self-rendered; we make no claims about real clients) and training a perception head with dense tile-level supervision plus rule-based assembly reproduces \emph{all} encoder-input planes exactly for 92.1\% of test states; feeding the perceived state through the frozen \emph{Transformer} encoder of Table~\ref{tab:offline} and the rationale LLM yields 92.7\% \mortal{}-agreement end-to-end---1.96 points below the same encoder on clean states (94.7\%), i.e.\ the perception head's error budget. Two construction notes: this e2e path regenerates the fact string (including the ranking) from the \emph{perceived} state, so the action equals the encoder's argmax on the perceived state by construction and 92.7\% measures perception fidelity, not LLM skill; the LLM's role in this path is the rationale text (90.7\% sentences exact). The comparison baseline is a \emph{trained} VLM: Qwen3.5-0.8B's native vision tower (0.1B visual parameters) with LoRA, trained on the same 100k human-labeled decisions, acting directly from the image: 68.6\% (the untrained VLM never produces valid JSON). The 24-point gap is confounded with both training-data scale (1.71M vs.\ 100k decisions) and supervision (the encoder distills \mortal{}, the VLM imitates human records), so it measures this pipeline's division of labor, not VLMs in general.

\section{Replication Notes on the Original Pipeline}
\label{sec:replnotes}

\paragraph{The ``winner filter'': a paper--code mismatch.} The paper states that only the game winner's decisions are kept; the released code in fact keeps, within every hand (kyoku), the decisions of the player ranked first \emph{by score at the end of that hand}. Both rules are outcome-conditioned data selection, a standard filtered-behavior-cloning practice. The mismatch still matters for replication: what is replicated is the code's rule (which we follow for comparability), not the paper's description, and we did not ablate either filter against no filter. We note the filter's purpose is unclear in the original even for \mortal{}-labeled training, where the labels come from the teacher rather than the filtered winner.

\paragraph{Private patches.} Both the observation API and the LLM-in-arena engine hook described in \S\ref{sec:setup} are absent from the public repository; any independent replication must rebuild them, which we release.

\section{Negative Results and Practical Pitfalls}
\label{sec:negative}

Each item below cost hours and is documented with logs:
(1) a policy-KL regularizer routed through a frozen encoder trained on 0/1 inputs \emph{harms} perception training (plane BCE 0.22 vs.\ 0.135 without);
(2) DETR-style query-token decoders fail to learn tile reading within 1{,}000 steps, while a simple grid convolution works;
(3) \texttt{AutoModelForCausalLM} silently misloads a hybrid-attention multimodal model so that LoRA patches nothing (first token reverts to \texttt{<think>}); loading via \texttt{AutoModelForImageTextToText} is required---the bug was caught only because the patched model's logits came out \emph{identical} to the base's;
(4) \texttt{label\_smoothing} in cross-entropy spreads probability over \emph{illegal} actions masked to $-10^4$, exploding the loss to $300+$;
(5) fused liger losses do not apply at validation time, doubling peak memory (7.4$\to$15.6\,GB) and crashing 2B training at step 400;
(6) an evaluation harness must self-check: our first strength evaluation returned four games with placement distribution $[0,0,0,0]$ and ``average placement 50.0''---a parsing bug that a sanity assertion catches immediately.

\section{Limitations and Ethics}
\label{sec:limits}

\textbf{Data and licensing.} Game records are from Tenhou's paid phoenix tables; their terms (\url{https://tenhou.net/man/}, accessed 2026-10-25) do not expressly authorize redistribution of derived artifacts, so withholding weights is our conservative reading, not a claimed legal necessity: \emph{we release no trained weights} (LoRA adapters, encoders, perception modules)---only code, configurations, evaluation protocols, and results. We are contacting Tenhou before posting this preprint and will report their response. The reconstructed \texttt{libriichi} patches are derivative works of the AGPL-3.0 engine, so the released harness inherits AGPL-3.0; reproducing the arena additionally requires the user to supply a \mortal{} checkpoint such as VoidShine-298k, which we do not distribute.
\textbf{Scale.} Our claims are about $\leq$2B models; the original 7B--9B result is not re-evaluated here. Within the text-LLM family, the \emph{absence of a detectable} placement difference rests on 100 paired games per pair ($p{=}1.0$ / $0.37$). Across families, only the scaled encoder separates from the LLMs significantly---and it differs in data (17$\times$), labels (distilled), and architecture simultaneously; the matched-data CNN-vs-LLM gap is not significant. Single 100-game blocks have $\approx$0.10-rank s.e.\ each, so differences of independent blocks carry an s.d.\ of $\approx$0.14 ranks---we report both blocks for the \mortal{}-labeled 0.8B.
\textbf{Supervision detail.} The two scaled encoders also distill \mortal{}'s Q-distribution; a hard-label control (\S\ref{sec:offline}) bounds this effect, and the matched-data CNN-vs-LLM comparison is hard-vs-hard throughout.
\textbf{Opponent.} Arena placements are against a specific community \mortal{} checkpoint, not the original paper's (unreleased) \mortal{}; absolute placements are not directly comparable to the paper's figure.
\textbf{Vision.} The pixel pipeline uses self-rendered boards, not real client screenshots.
\textbf{Seeds.} One training run per configuration; one greedy decoding pass; arena SEs reflect game sampling only.

\section{Conclusion}
\label{sec:conclusion}

On one consumer GPU, the mahjong conclusions of \citet{llm4cardgame} shrink but do not collapse: small text-prompt LLMs do learn to play (placements far above chance), yet a two-order-of-magnitude smaller structured encoder beats them at imitation under matched data and labels, teacher labels raise teacher agreement more than a 2.5$\times$ parameter increase does, and---once the prescribed game-play evaluation is actually run---few-point agreement gains within the LLM family do not translate into a detectable placement difference at 100-game power, while the scaled encoder reaches parity with the champions. Notably, the original paper's 7--9B models ($\approx$2.6 average placement against their own \mortal{}, metric-comparable to ours) sit between our encoders (2.51) and our LLMs (2.95--3.03), so whether text is a bottleneck at their scale is a genuinely open question---one this budget could not test. The constructive lesson is division of labor: let structure carry decisions, let programs carry facts, and keep the language model for what it demonstrably does here---faithfully integrating program-supplied facts into readable rationales. We release everything we can within Tenhou's terms.

\paragraph{Reproducibility statement.} Code, configurations, the reconstructed engine patches, all $\approx$3{,}230 arena game logs, per-run logs with line-numbered provenance for every number in this paper, and the full internal audit trail are included in the artifact. Training requires only the public code plus a licensed copy of the data.

\bibliographystyle{plainnat}
\bibliography{refs}

\appendix

\section{Hyperparameters and Run Inventory}
\label{app:hparams}

\paragraph{Board encoders.} Input is a $58\times34$ uint8 plane stack + 71 scalars computed deterministically from the 36 prompt fields. CNN: Conv1d $58{\to}192$ + positional embedding + FiLM conditioning on the scalars + 10 residual blocks (Conv1d--GroupNorm--SE attention), policy head over 46 actions; 2.61M parameters. Transformer: one token per tile kind (34) + one context token, 6 encoder layers, width 256, 8 heads, pre-norm; 4.85M parameters. Both: AdamW, lr $10^{-3}$, weight decay 0.05, batch 1024, up to 16 epochs, early stopping on a per-game 3\% validation split; label smoothing 0.05; ``soft'' adds a CE term to $\mathrm{softmax}(Q/T)$ with $T{=}1$, $\alpha{=}0.5$ (plus a value head for the TF row). The matched-data CNN (100k) trains in 1.6\,min; the TF on 1.71M decisions in 28\,min.

\begin{table}[h]
\centering\small
\caption{Common LLM SFT configuration (LLaMA-Factory).}
\begin{tabular}{ll}
\toprule
LoRA rank / target & 8 (default) / all linear layers \\
template & \texttt{qwen3\_5\_nothink} (assistant answers directly; no think block) \\
epochs / steps & 1 / 6{,}188 (100k decisions, effective batch 16) \\
optimizer / lr & AdamW / $10^{-4}$ cosine, 10\% warmup \\
length / precision & 4{,}096 tokens / bf16 + liger kernels \citep{liger} \\
\bottomrule
\end{tabular}
\end{table}

Training wall-clock on the single RTX 5070 Ti: 0.8B 9.8\,h; 0.8B (\mortal{} labels) 10.1\,h; 2B 13.1\,h (one run rented an A100-40GB for \$21); encoders 1.6--33 min; perception v5 175 min; the tri-modal (pixels$+$tokens$+$facts) SFT 307 min; each arena campaign $\approx$1.5--4\,h per 100 games.

\section{Arena Protocol Details}
\label{app:arena}
Challenger occupies one seat per game; 25 seeds $\times$ 4 seat rotations $=$ 100 games per agent, game seed constant. LLM agents ran on disjoint 100-game seed blocks (92{,}000/93{,}000/94{,}000, plus the \mortal{}-labeled 0.8B rerun on 92{,}000 and the \mortal{}-labeled 2B's first arena run on 94{,}000, both for paired tests); the encoders additionally ran 100-game blocks on 91{,}000 and 92{,}000 (exactly paired with each other and with the 0.8B rerun) and 1{,}000-game runs on 95{,}000--95{,}249, plus two extra-opponent matches (vs.\ a pool of three matched-data CNN encoders, and vs.\ three copies of our own distilled \mortal{}-derivative checkpoint). s.e.'s for the 1{,}000-game runs are clustered by seed (4 games per seed). Run inventory: random floor; 0.8B-human (92k); 0.8B-\mortal{} A/B (93k/92k); 2B-human and 2B-\mortal{} (94k, paired); encoders TF/CNN on 91k, 92k, and 95k$\times$10; TF vs.\ encoder pool; TF vs.\ distilled \mortal{}; 32-game \mortal{} self-symmetry. Greedy decoding throughout matches the original protocol's configuration. Decoding: greedy, \texttt{max\_new\_tokens}=32, chat template identical to training; encoders take a single argmax over the engine's legal mask. Unparseable or illegal LLM outputs would fall back to a random legal action (the original engine's behavior); in practice the fallback never triggered (format validity 100\%). Placement points follow the original arena ($90/45/0/-135$). The random-legal floor validates the harness end to end; a Mortal-vs-Mortal check ($[8,8,8,8]$ over 32 games) additionally validates the bookkeeping (identical deterministic agents under seat rotation are exchangeable by construction; \mortal{} here plays greedily, as in all our matches). Behavior statistics (deal-in, riichi, meld, win rates) are computed from the mjai game logs.

\section{Number Provenance}
Every number in this paper maps to a line in a log file in the artifact; the mapping table is shipped as \texttt{NUMBERS.md} (41 audited entries covering every quantitative claim; further intermediate numbers trace to the same logs) alongside a full internal audit report that re-verified $\sim$150 citations against raw logs.

\end{document}